\documentclass[runningheads]{llncs}
\usepackage{tcolorbox}
\usepackage{graphicx}
\usepackage{amsmath}
\usepackage{soul}
\usepackage{float}
\usepackage{xcolor}

\usepackage[inline]{enumitem}
\begin{document}
\title{Identification of COVID-19 related Fake News via Neural Stacking}
%
%
\author{Boshko Koloski\inst{1,2} \and
Timen Stepi\v{s}nik-Perdih\inst{3} \and
Senja Pollak\inst{1} \and \\
Bla\v{z} \v{S}krlj\inst{1,2} }
\authorrunning{Koloski et al.}
\institute{Jo\v{z}ef Stefan Institute, Jamova 39, 1000 Ljubljana, Slovenia \and
Jo\v{z}ef Stefan Int. Postgraduate School, Jamova 39, 1000 Ljubljana, Slovenia \and
University of Ljubljana, Faculty of Computer and Information Science, Ve\v{c}na pot 113, Ljubljana, Slovenia \\
\email{\{boshko.koloski,blaz.skrlj\}@ijs.si}}
\maketitle              
\begin{tcolorbox}

Please cite the original version, with the following doi: 10.1007/978-3-030-73696-5\_17 (at CONSTRAINT 2021: Combating Online Hostile Posts in Regional Languages during Emergency Situation pp. 177-188) and is available online at \url{https://link.springer.com/chapter/10.1007/978-3-030-73696-5\_17}.

\end{tcolorbox}

\begin{abstract}
 Identification of Fake News plays a prominent role in the ongoing pandemic, impacting multiple aspects of day-to-day life. In this work we present a solution to the shared task titled \textit{COVID19 Fake News Detection in English}, scoring the 50th
 place amongst 168 submissions. The solution was within 1.5\% of the best performing solution. The proposed solution employs a heterogeneous representation ensemble, adapted for the classification task via an additional neural classification head comprised of multiple hidden layers.  The paper consists of detailed ablation studies further displaying the proposed method's behavior and possible implications. The solution is freely available. \\ \url{https://gitlab.com/boshko.koloski/covid19-fake-news}
\end{abstract}

\section{Introduction}

Fake news can have devastating impact on the society. 
In the times of a pandemic, each piece of information can have a significant role in the lives of everyone. The verification of the truthfulness of a given information as a fake or real is crucial, and can be to some extent learned~\cite{patwa2021overview}.
Computers, in order to be able to solve this task, need the data represented in a numeric format in order to draw patterns and decisions. We propose a solution to this problem by employing various natural language processing and learning techniques. 
\par The remainder of this work is structured as follows: Section \ref{sec:related_work} describes the prior work in the field of detection of fake-news. The provided data is described in Section \ref{sec:data} and Section \ref{sec:representaitons} explains our proposed problem representation approaches while Section \ref{sec:meta_models} introduces two different meta-models built on top of the basic representations listed in Section \ref{sec:representaitons}. The experiments and results achieved are listed in Section \ref{sec:expr}, finally the conclusion and the proposed future work are listed in Section \ref{sec:final_work}.

\section{Related Work}
\label{sec:related_work}

    The fake-news text classification task \cite{shannon2003cytoscape} is  defined as follows: given a text and  a set of possible classes \textit{fake} and \textit{real}, to which a text can belong, an algorithm is asked to predict the correct class of the text. Most frequently, fake-news text classification refers to classification of data from social media. 
    The early proposed solutions to this problem used hand crafted features of the authors such as word and character feature distributions. Interactions between fake and real news spread on social media gave the problem of fake-news detection a network-alike nature\cite{shu2019studying}. The network based modeling discovered useful components of the fake-news spreading mechanism and led to the idea of the detection of bot accounts \cite{shao2018spread}.

    Most of the current state-of-the-art approaches for text classification leverage large pre-trained models like the one Devlin et al. \cite{devlin2018bert} and have promising results for detection of fake news \cite{jwa2019exbake}. However for fake-news identification tasks, approaches that make use of n-grams and the Latent Semantic Analysis \cite{lsa} proved to provide successful solutions on this task (see Koloski et al. \cite{koloski2020multilingual}). Further enrichment of text representations with taxonomies and knowledge graphs\cite{SKRLJ2020101104} promises improvements in performance.
    
\section{Data description}
\label{sec:data}
In this paper we present a solution to the subset of the fake-news detection problem - The identification of COVID-19 related Fake News \cite{patwa2020fighting,patwa2021overview}. The dataset consists of social media posts in English collected from Facebook, Twitter and Instagram, and the task is to determine for a given post if it was real or fake in relation to COVID-19. The provided dataset is split in three parts: train, validation and test data. The distribution of data in each of the data sets is shown in Table \ref{tab:data_dist}.
\begin{table}[H]
    \centering
    \caption{Distribution of the labels in all of the data splits.}
    \begin{tabular}{c|c|c|c}
        part & train & validation & test  \\ \hline
        size & 6420 & 2140 & 2140 \\ \hline
        real & 3360(52\%) & 1120(52\%) & 1120(52\%)   \\ \hline
        fake & 3060(48\%) & 1020(48\%) & 1020(48\%) \\
    \end{tabular}
    \label{tab:data_dist}
 
\end{table}
  
\section{Proposed method}
The proposed method consists of multiple submethods that aim to tackle different aspects of the problem. On one side we focus on learning the hand crafted features of authors and on the other we focus on learning the representation of the problem space with different methods. 
\label{sec:representaitons}
\subsection{Hand crafted features}
\subsubsection{Word based}
Maximum and minimum word length in a tweet, average word length, standard deviation of the word length in tweet. Additionally we counted the number of words beginning with upper and the number of words beginning a lower case.
\subsubsection{Char based}
The character based features consisted of the counts of digits, letters, spaces, punctuation, hashtags and each vowel, respectively.

\subsection{Latent Semantic Analysis}
    Similarly to Koloski et al. \cite{koloski2020multilingual} solution to the PAN2020-Fake News profiling we applied the low dimensional space estimation technique. First we preprocessed the data by lower-casing the tweet content and removing the hashtags and punctuation. After that we removed the stopwords and obtained the final clean presentation. From the cleaned text, we generated the POS-tags using the NLTK library\cite{nltk}.
    
    \begin{figure}[h]
        \centering
        \resizebox{\columnwidth}{!}{\includegraphics{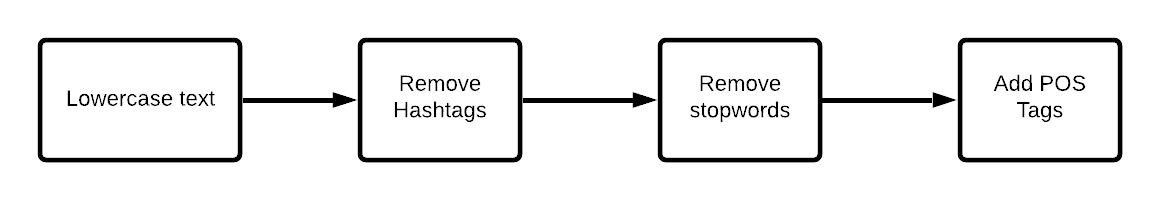}}
        \caption{Text preparation for the LSA.}
        \label{fig:lsa_prep}
    \end{figure}
    For the feature construction space we used the technique used by Martinc et al. \cite{martincPan} which iteratievly weights and chooses the best n-grams.
    We used two types of n-grams:
    \begin{itemize}
        \item Word based: n-grams of size 1 and 2
        \item Character based: n-grams of size 1, 2 and 3
    \end{itemize}
    We generated $n$ features with $n/2$ of them being word and $n/2$ character n-grams. We calculated TF-IDF on them and preformed SVD \cite{halko2009finding} 
    With the last step we obtained the LSA representation of the tweets. 

\begin{figure}[H]
    \centering
    \resizebox{\columnwidth}{!}{\includegraphics{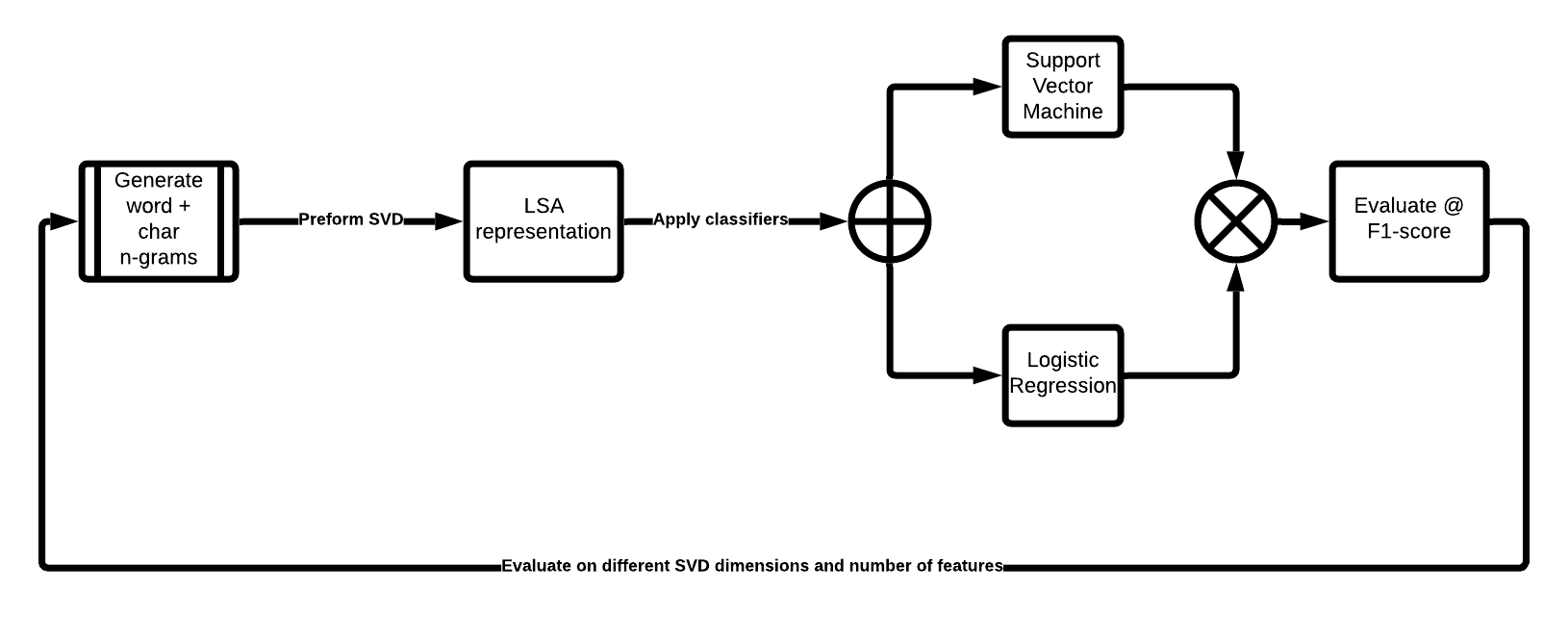}}
    \caption{The proposed learning procedure with the LSA. The evaluation is performed on validation dataset.}
    \label{fig:lsa_pipeline}
    \end{figure}

    For choosing the optimal number of features \textbf{n} and number of  dimensions \textbf{d}, we created custom grid consisted of $n` \in [500,1250,2500,5000,10000,15000,20000]$ and $d` \in [64,128,256,512,768]$. 
    For each tuple $(n`,d`) , n` \in $ \textbf{d}  and $ d` \in $ \textbf{d} we generated a representation and trained (SciKit library \cite{scikit-learn}) \textit{SVM} and a \textit{LR} (Logistic regression) classifier. The learning procedure is shown in Figure \ref{fig:lsa_pipeline}. 
    
    The best performing model had $2500$ features reduced to $512$ dimensions. 
    
\subsection{Contextual features}
We explored two different contextual feature embedding methods that rely on the transformer architecture. The first method uses the already pretrained \textit{sentence\_transfomers} and embedds the texts in an unsupervised manner. The second method uses DistilBERT which we fine tune to our specific task.
\subsubsection{sentence\_transfomers}
    For fast document embedding we used three different contextual embedding methods from the \textit{sentence\_transfomers} library \cite{reimers-2019-sentence-bert}: 
    \begin{itemize}
        \item \textit{distilbert-base-nli-mean-tokens}
        \item \textit{roberta-large-nli-stsb-mean-tokens}
        \item \textit{xlm-r-large-en-ko-nli-ststb}
    \end{itemize}
    First, we applied the same preprocessing as shown in Figure \ref{fig:lsa_prep}, where we only excluded the POS tagging step. 
    After we obtained the preprocessed texts we embedded every tweet with a given model and obtained the vector representation. 
    After we obtained each representation, we learned a Stochastic Gradient Descent based learner, penalizing both the "linear" and "hinge" loss parameters. The parameters were optimized on a GridSearch with a 10-fold Cross-validation on every tuple of parameters.  
\subsubsection{DistilBERT}
is a distilled version of BERT that retains best practices for training BERT models \cite{dbert}. It is trained on a concatenation of English Wikipedia and Toronto Book Corpus. To produce even better results, we fine-tuned the model on train data provided by the organizers. BERT has its own text tokenizer and is not compatible with other tokenizers so that is what we used to prepare data for training and classification.

\subsection{tax2vec features}


 tax2vec \cite{SKRLJ2020101104} is a data enrichment approach that constructs semantic features useful for learning. It leverages background knowledge in the form of taxonomy or knowledge graph and incorporates it into textual data. We added generated semantic features using one of the two approaches described below to top 10000 word features according to the TF-IDF measure.
We then trained a number of classifiers on this set of enriched features (Gradient boosting, Random forest, Logistic regression and Stochastic gradient descent) and chose the best one according to the F1-score calculated on the validation set..
\textbf{Taxonomy based (tax2vec). }
Words from documents are mapped to terms of the WordNet taxonomy \cite{wn}, creating a document-specific taxonomy after which a term-weighting scheme is used for feature construction. Next, a feature selection approach is used to reduce the number of features.
\textbf{Knowledge Graph based (tax2vec(kg)). }
Nouns in sentences are extracted with SpaCy and generalized using the Microsoft Concept Graph \cite{article} by "is\_a" concept. A feature selection approach is used to reduce the number of features.

\section{Meta models}
\label{sec:meta_models}
From the base models listed in Section \ref{sec:representaitons} we constructed two additional meta-models by combining the previously discussed models.

\subsection{Neural stacking}  
In this approach we learn a dense representation with 5-layer deep neural network. For the inputs we use the following representations: 
\begin{itemize}
    \item LSA representation with $N = 2500$ features reduced to $d = 256$ dimensions.
    \item Hand crafted features - $d = 16$ dimensions
    \item \textit{distilbert-base-nli-mean-tokens} - $d = 768$ dimensions
    \item \textit{roberta-large-nli-stsb-mean-tokens} - $d = 768$ dimensions
    \item \textit{xlm-r-large-en-ko-nli-ststb} - $d = 768$ dimensions
\end{itemize}
This represents the final input $X_{N x 2576}$ for the neural network. After concatenating the representations we normalized them. 
We constructed a custom grid consisted of learning\_rate = $\lambda \in$  [0.0001, 0.005, 0.001, 0.005, 0.01, 0.05, 0.1], dropout = $p \in$ [0.1, 0.3, 0.5, 0.7],
batch\_size $\in$ [16,32,64,128,256],
epochs $\in$ [10, 100, 1000]. In the best configuration we used the \textit{SELU} activation function and dropout $p = 0.7$ and learning rate $\lambda = 0.001$. The loss function was \textit{Cross-Entropy} optimized with the $Stochastic Gradient Optimizer$, trained on $epochs = 100$ and with $batch\_size = 32$. \\
Layers were composed as following: 
\begin{itemize}
    \item \textit{input} layer - $d = 2576$ nodes
    \item $dense_{1}$ layer - $d = 896$ nodes, activation = \textit{SELU}
    \item $dense_{2}$ layer - $d = 640$ nodes, activation = \textit{SELU} 
    \item $dense_{3}$ layer - $d = 512$ nodes, activation = \textit{SELU}
    \item $dense_{4}$ layer - $d = 216$ nodes, activation = \textit{SELU} 
    \item $dense_{5}$ layer - $d = 2$   nodes, activation = \textit{Sigmoid}
\end{itemize}

\subsection{Linear stacking}
    The second approach for meta-learning considered the use of the predictions via simpler models as the input space. We tried two separate methods:
    \subsubsection{Final predictions}
\begin{figure}[H]
        \centering
        \includegraphics[width=10cm,height=6.75cm,keepaspectratio]{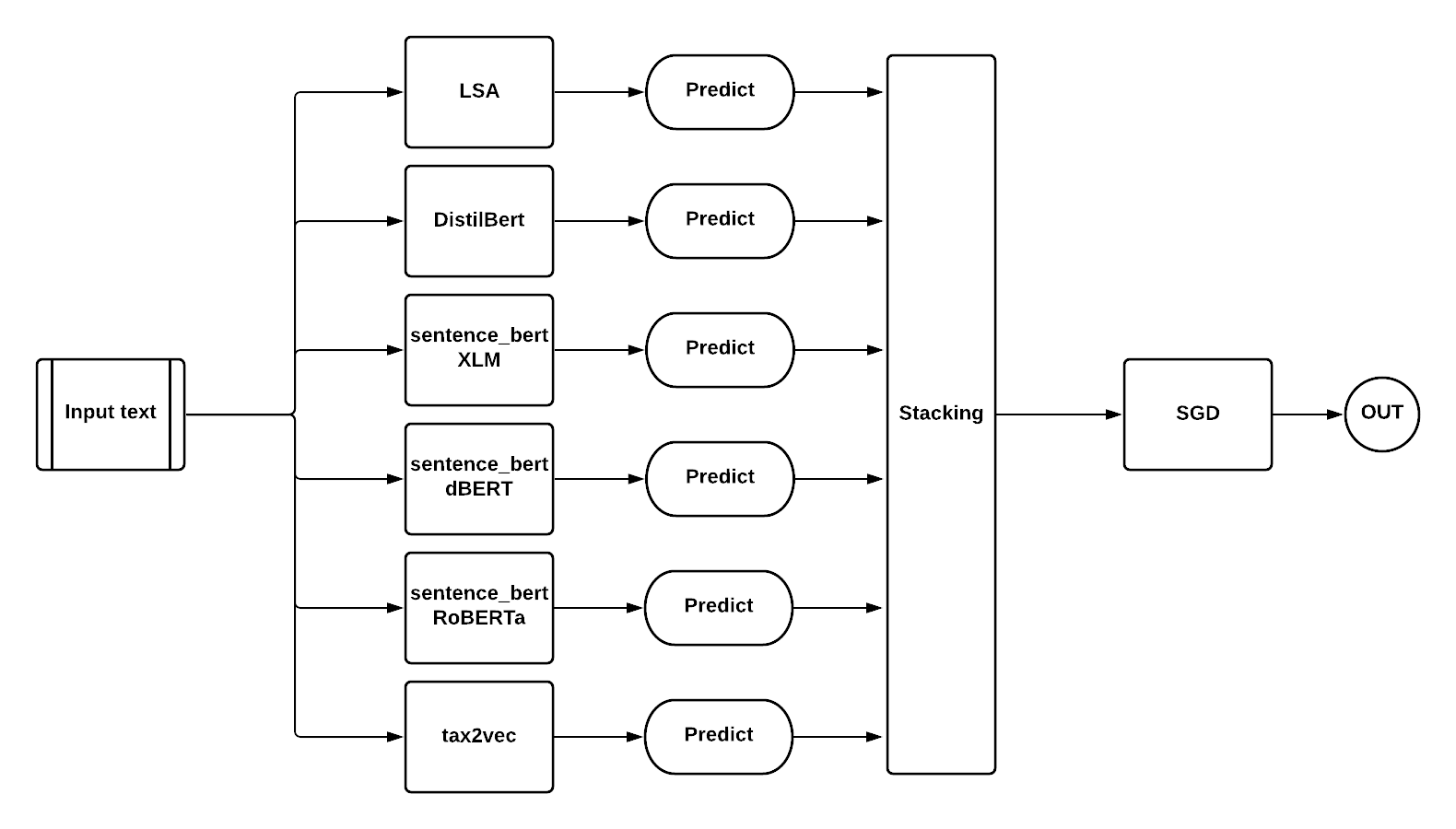}
        \caption{Stacking architecture based on base model predictions.}
        \label{fig:stacking_preds}
    \end{figure}

        We considered the predictions from the \textit{LSA}, \textit{DistilBert}, \textit{dbert}, \textit{xlm}, \textit{roberta}, \textit{tax2vec} as the input. From the models' outputs we learned a Stochastic Gradient Optimizer on 10-fold CV. The learning configuration is shown in Figure \ref{fig:stacking_preds}. 
    \subsubsection{Decision function-based prediction}
        In this approach we took the given classifier's value of the decision function as the input in the stacking vector. For the SVM based SGD we used the \textit{decision\_function} and for the Logistic Regression we used the \textit{Sidmoid\_activation}. The proposed architecture is similar to the architecture in Figure \ref{fig:stacking_preds}, where prediction values are replaced by decision function values. 
    
\section{Experiments and results}
\label{sec:expr}
This section describes model parameters, our experiments and the results of experiments as well as the results of the final submission. 

We conducted the experiments in two phases. The experiment phases synced with the competition phases and were defined as \textit{TDT} phase and \textit{CV} phase. In the TDT phase the train and validation data 
is split into three subsets, while in the CV phase all data 
is concatenated and evaluated on 10-folds.

\begin{table}[H]
    \centering
    \caption{Final chosen parameters for the best model of each vectorization.} 
    \resizebox{\textwidth}{!}{\begin{tabular}{|c|c|c|}
        \hline
        Vectorization & Model & Parameters\\ \hline
        LSA &  LR & \textit{'l1\_ratio': 0.05, 'penalty': 'elasticnet', 'power\_t': 0.5} \\ \hline
        Hand crafted features  & SVM & \textit{ 'l1\_ratio': 0.95, 'penalty': 'elasticnet', 'power\_t': 0.1} \\ \hline
        distilbert-base-nli-mean-tokens & LR & 'C': 0.1, 'penalty':'l2'  \\ \hline
        roberta-large-nli-stsb-mean-tokens & LR & 'C':'0.01', 'penalty': 'l2' \\ \hline
        xlm-r-large-en-ko-nli-ststb  & SVM & 'C': 0.1, 'penalty': 'l2'\\ \hline
        linear stacking\_probs & SGD  &  'l1\_ratio': 0.8, 'loss': 'hinge', 'penalty': 'elasticnet'  \\ \hline
        linear stacking  & SGD & 'l1\_ratio': 0.3, 'loss': 'hinge', 'penalty': 'elasticnet'\\ \hline
        tax2vec\_tfidf  & SGD & 'alpha': 0.0001, 'l1\_ratio': 0.15,  'loss': 'hinge', 'power\_t': 0.5 \\ \hline
        tax2vec(kg)\_tfidf & SVM  & 'C': 1.0,  'kernel': 'rbf'    \\ \hline
    \end{tabular}}
    \label{tab:params_TDT}
\end{table}

\subsection{Train-development-test (TDT) split}
\begin{figure}[!ht]
    \includegraphics[width=\textwidth]{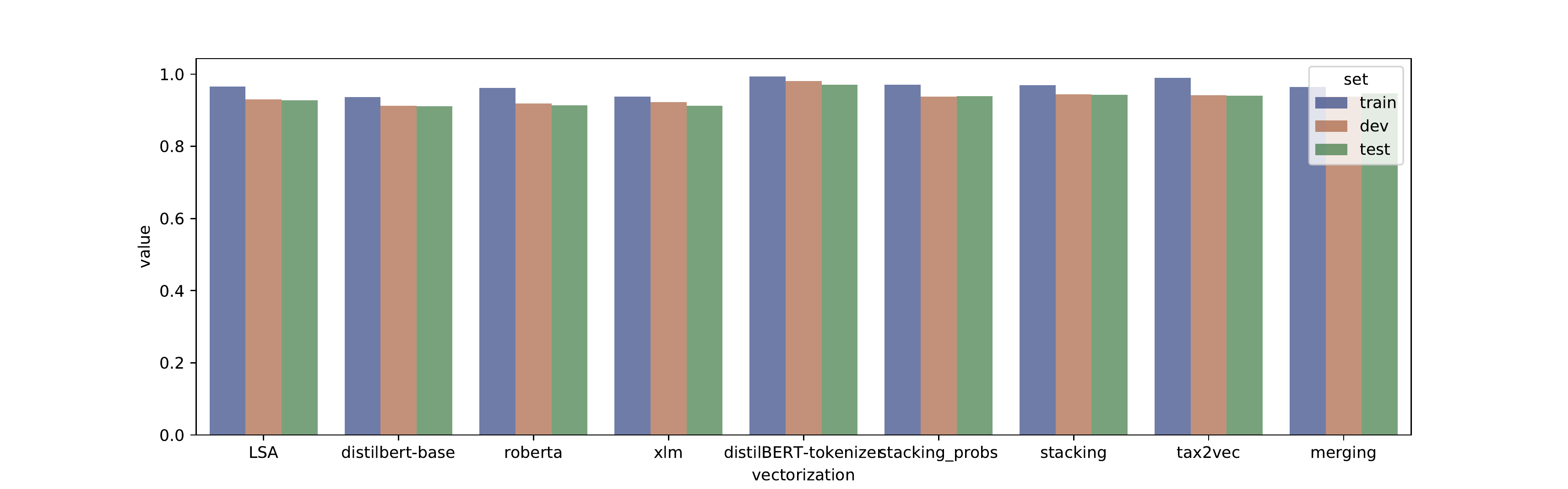}
    \caption{Best performing methods of feature vectorization according to F1-score.}
    \label{tab:tdt_evaluation_figure}
\end{figure}

In the first phase, we concatenated the train and the validation data and splitted it into three subsets: 
\textit{train}(75\%), \textit{dev}(18.75\%) and \textit{test}(6.25\%). On the \textit{train} split we learned the classifier which we validated on the \textit{dev} set with measurement of F1-score. Best performing model on the \textit{dev} set was finally evaluated on the \text{test} set. 
Achieved performance is presented in Table \ref{tab:tdt_evaluation} and the best performances are shown in Figure \ref{tab:tdt_evaluation_figure}.

\begin{table}[H]
    \centering
    \caption{F1-scores for different methods of vectorization on the TDT data split.}
    \resizebox{\columnwidth}{!}{\begin{tabular}{|c|c|c|c|c|}
    \hline
    Vectorization & Train F1-score & DEV F1-score &  Test F1-score \\ \hline
    distilBERT-tokenizer   & \textbf{0.9933} & \textbf{0.9807} & \textbf{0.9708} \\ \hline
     neural stacking & 0.9645 & 0.9377 & 0.9461 \\ \hline
     linear stacking  & 0.9695 & 0.9445 & 0.9425 \\ \hline
     tax2vec  & 0.9895  & 0.9415 & 0.9407  \\ \hline
    linear stacking\_probs& 0.9710 & 0.9380 & 0.9390 \\ \hline
     LSA&  0.9658 & 0.9302 & 0.9281 \\ \hline
      roberta-large-nli-stsb-mean-tokens & 0.9623 & 0.9184 & 0.9142 \\ \hline
     xlm-r-large-en-ko-nli-ststb  &0.9376 & 0.9226 & 0.9124 \\ \hline
     distilbert-base-nli-mean-tokens & 0.9365 & 0.9124 & 0.9113 \\ \hline
    
     tax2vec(kg) & 0.8830  & 0.8842 & 0.8892 \\ \hline
    Hand crafted features  & 0.7861 & 0.7903 & 0.7805 \\ \hline
     
    \end{tabular}}
    \label{tab:tdt_evaluation}
\end{table}

DistilBERT comes out on top in F1-score evaluation on all data sets in TDT data split---to the extent that we feared overfitting on the train data---while handcrafting features did not prove to be successful.
Taxonomy based tax2vec feature construction trails distilBERTs score but using a knowledge graph to generalize constructed features seemed to decrease performance significantly (tax2vec(kg)).
Other methods scored well, giving us plenty of reasonably good approaches to consider for the CV phase.

\subsection{CV split}
In the second phase - the \textit{CV} phase we concatenated the data provided by the organizers and trained models on 10-fold Cross-Validation. The evaluation of the best-performing models is presented in Table \ref{tab:cv_train}.

During cross-validation, LSA showed consistency in good performance. With similar performance were the tax2vec methods which this time scored very similarly.

\begin{table}[H]
    \centering
        \caption{F1-scores of models when training using 10-fold cross-validation.}
        \begin{tabular}{|c|c|c|}
        \hline
        Model name & Vectorization & 10-fold CV \\ \hline
        LSA & LSA & \textbf{0.9436} \\ \hline
        sentence\_transformers & distilbert  & 0.9071 \\ \hline
        sentence\_transformers & roberta-large & 0.9077 \\ \hline
        sentence\_transformers & xlm-roberta & 0.9123 \\ \hline
        gradient boosting & tax2vec & 0.9335 \\ \hline
        gradient boosting & tax2vec(kg) & 0.9350 \\ \hline    \end{tabular}

    \label{tab:cv_train}
\end{table}

\subsection{Evaluating word features}
To better understand the dataset and trained models we evaluated word features with different metrics to pinpoint features with the highest contribution to classification or highest variance.

\subsubsection{Features with the highest variance}
We evaluated word features within the train dataset based on variance in \emph{fake} and \emph{real} classes and found the following features to have the highest variance: \\

\emph{"Fake" class}
\begin{itemize*}
  \item cure
  \item coronavirus
  \item video
  \item president
  \item covid
  \item vaccine
  \item trump
  \item 19
\end{itemize*}

\emph{"Real" class}
\begin{itemize*}
  \item number
  \item total
  \item new
  \item tests
  \item deaths
  \item states
  \item confirmed
  \item cases
  \item reported
\end{itemize*}

\subsubsection{SHAP extracted features}
After training the models we also used Shapley Additive Explanations \cite{NIPS2017_7062} to extract the most important word features for classification into each class. The following are results for the gradient boosting model: \\

\emph{"Fake" class}
\begin{itemize*}
  \item video
  \item today
  \item year
  \item deployment
  \item trump
  \item hypertext transfer protocol
\end{itemize*}

\emph{"Real" class}
\begin{itemize*}
  \item https
  \item covid19
  \item invoking
  \item laboratories
  \item cases
  \item coronavirus
\end{itemize*}

\subsubsection{Generalized features}
We then used WordNet with a generalizing approach called ReEx (Reasoning with Explanations)\footnote{https://github.com/OpaqueRelease/ReEx} to generalize the terms via the ``is\_a'' relation into the following terms: \\

\emph{"Fake" class}
\begin{itemize*}
  \item visual communication
  \item act
  \item matter
  \item relation
  \item measure
  \item hypertext transfer protocol
  \item attribute
\end{itemize*}

\emph{"Real" class}
\begin{itemize*}
  \item physical entity
  \item message
  \item raise
  \item psychological feature
\end{itemize*}

\subsection{Results}
Results of the final submissions are shown in Table \ref{tab:final_submission}.

\begin{table}[H]
    \centering
     \caption{Final submissions F1-score results.}
    \begin{tabular}{c | c | c}
        submission name & model & F1-score\\ \hline
        btb\_e8\_4 & neural stacking & \textbf{0.9720} \\ \hline
        btb\_e8\_3 & LSA & 0.9416 \\ \hline
        btb\_e8\_1 & tax2vec & 0.9219 \\ \hline
        btb\_e8\_2 & linear stacking & 0.8464 \\ \hline
        btb\_e8\_5 & distilbert & 0.5059 
    \end{tabular}
    \label{tab:final_submission}
\end{table}

\begin{figure}[H]
\centering
\begin{tabular}{ccc}
{\includegraphics[width = .32\linewidth]{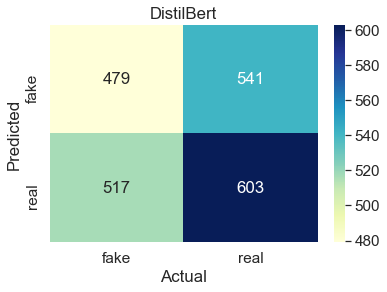}}
&
{\includegraphics[width = .32\linewidth]{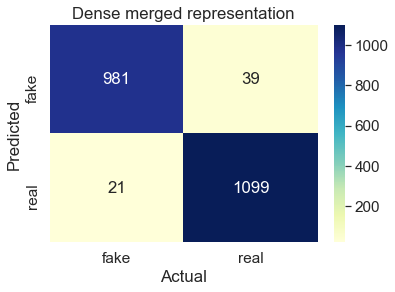}}
&
{\includegraphics[width = .32\linewidth]{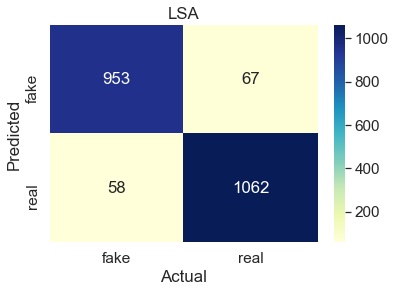}}
\\
{\includegraphics[width = .32\linewidth]{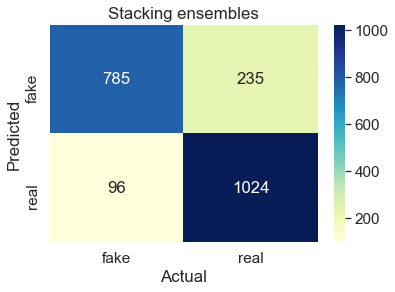}}
&
{\includegraphics[width = .32\linewidth]{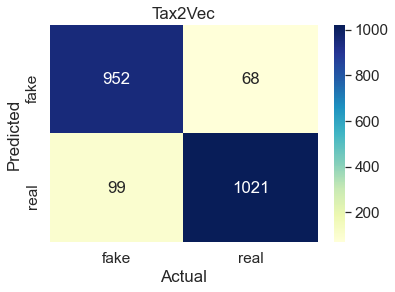}}
\\
\end{tabular}
\caption{Heatmaps of predicted and actual labels on final submission results.}
\label{fig:heat}
\end{figure}

DistilBERT appears to have overfitted the train data on which it achieved very high F1-scores, but failed to perform well on the test data in the final submission.
Our stacking method also failed to achieve high results
 in the final submission, being prone to predict ``fake" news as can be seen in Figure \ref{fig:heat}.
On the other hand, the taxonomy based tax2vec data enrichment method as well as the LSA model have both shown good results in the final submission, while our best performing model used stacking, where we merged different neural and non-neural feature sets into a novel representation. With this merged model, we achieved 0.972 F1-score and ranked 50th out of 168 submissions.

In Figure \ref{fig:heat} we present the confusion matrices of the models evaluated in the final submissions.

\section{Conclusion and further work}
 \label{sec:final_work}
    In our take to tackle the detection of fake-news problems we have have exploited different approaches and techniques. We constructed hand crafted features that captured the statistical distribution of words and characters across the tweets.
    From the collection of n-grams of both character and word-based features to be found in the tweets we learned a latent space representation, potentially capturing relevant  patterns. With the employment of multiple BERT-based representations we captured the contextual information and the differences between fake and real COVID-19 news. However such learning showed that even though it can have excellent 
    results for other tasks, for tasks such as classification of short texts it proved to fall 
    behind some more sophisticated methods. 
    To overcome such pitfalls we constructed two different meta models, learned from the decisions of simpler models. The second model learned a new space from the document space representations of the simpler models by embedding it via a 5 layer neural network. This new space resulted in a very convincing 
    representation of this problem space achieving F1-score of \textbf{0.9720} on the final (hidden) test set. 
    \par For the further work we suggest improvements of our methods by the inclusion of background knowledge to the representations in order to gain more instance separable representations. We propose exploring the possibility of adding model interpretability with some attention based mechanism. Finally, as another add-on we would like to explore how the interactions in the networks of fake-news affect our proposed model representation.

\section{Acknowledgements}
The work of the last author was funded by the Slovenian Research Agency (ARRS) through a young researcher grant.
The work of other authors was supported by the Slovenian Research Agency core research programme \emph{Knowledge Technologies} (P2-0103) and the ARRS funded research projects \emph{Semantic Data Mining for Linked Open Data (ERC Complementary Scheme, N2-0078)} and \emph{Computer-assisted multilingual news discourse analysis with contextual embeddings - J6-2581)} . The work was also supported by European Union's Horizon 2020 research and  innovation programme under grant agreement No 825153, project EMBEDDIA (Cross-Lingual Embeddings for
Less-Represented Languages in European News Media).

%
%
%
%







\bibliographystyle{splncs04}
\bibliography{refs}

\end{document}